# Robotic bees: Algorithms for collision detection and prevention

Vincent Arcila
Universidad EAFIT
Colombia
vaarcilal@eafit.edu.co

Isabel Piedrahita
Universidad EAFIT
Colombia
ipiedrahiv@eafit.edu.co

Mauricio Toro
Universidad EAFIT
Colombia
mtorobe@eafit.edu.co

**ABSTRACT.** In the following paper we will discuss data structures suited for distance threshold queries keeping in mind real life application such as collision detection on robotic bees. We will focus on spatial hashes designed to store 3D points and capable of fastly determining which of them surpass a specific threshold from any other. In this paper we will discuss related literature, explain in depth the data structure chosen with its design criteria, operations and speed and memory efficiency analysis.

## 1. INTRODUCTION

In recent years the population of bees has been constantly decreasing . More than a billion bees have died in Colombia alone in the span of the last four years[1], worldwide, bees are not doing any better. With the number of bees dwindling dramatically in the past 10 years it is time to consider the importance of bees in the ecosystem, and consider ways to replace them in their duties. Robotic bees capable of visiting and pollinating multiple flowers in a day, could be a viable solution. In order to prevent these hypothetical robotic bees from colliding, we intend to keep them at no less than one-hundred meters of distance from one another by implementing a spatial hashing data structure capable of carrying out said proximity queries.

## 2. PROBLEM

Given a set of N points, which will represent the swarm of bees, in the 3D space return a set with the points that are closer than a threshold X in meters from any other point. By doing this we can alert these proposed robotic bees of any risk of collision so that their routes can be altered accordingly, keeping the bees functional for a longer time, and saving money in the long run, as they will not have to be replaced. In this paper we arbitrarily chose 100 meters as X, but our approach will maintain its characteristics regardless of the value of X.

## 3. RELATED WORK

### 3.1 Better collisions and faster cloth for Pixar's Coco[4]

While making Pixar's Coco animators ran into a complication, it was nearly impossible to get the cloth of clothes to sit properly on the huge cast of esqueletal characters, whose bones pinched the fabric causing what producers referred to as "Skeleton Wedgies". Because of this, their in-house cloth system, Fritz, had to implement a more efficient and easily systematized version of continuous collision detection.

In order to fix this, the Global Intersection Analysis, an algorithm that uses constant collision detection to identify when cloth has crossed over to the wrong side by analysing all the intersections between the digital mesh that composes the costumes. This was used to create a fast way to stop the cloth from going through and getting stuck between the small geometries that make up Coco's skeletons.

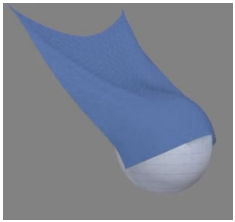

*Fig 1. Simulates sphere under cloth*

### 3.2: Search algorithms for swarms of robots: A survey[5]

This paper serves as an introduction to any scientist interested on understanding ways to keep an eye on groups of robots and to make them work more efficiently. It covers various algorithmic related problems in order to give the reader an idea of what to have in mind when designing systems that involves groups of robots.

One of the problems that needs to be solved in software is the problem of target search and tracking, understanding a target as an object with which the robots need to interact directly in order to succeed on their task. Understanding how to handle with software all the different environments that the robots might have to face is not as simple as one might think. The number of targets may be unknown, or may vary with time. Also, how to manage robots so that they do not act on the same target simultaneously, known as the task allocation problem is a thing to have in mind. Targets might not be static, which has to be considered.

Cooperation and coordination among the robots is an essential aspect that has to be taken into account when designing swarm robots systems. An ideal system would be the one on which all robots are exchanging information, combining measurements from multiple positions to accurately determine the targets position, and get the best synchronization possible.

### 3.3: Real-Time Robot Motion Planning in Dynamic Environments[7]

Navigation is one of the main fields collision detection is used on. As soon as a robot can move on its own, it has to decide where to, and this is where collision detection comes in very handy. Although there has already been extensive study of collision detection, further study is very important, because current algorithms understand the world around them via a group of queries, instead of using geometrical analysis.

Because of this, the process of collision detection is usually divided in two phases.

"Broad-Phase: Determines which geometric objects need to be checked for a certain query, ruling out distant objects based on bounding boxes or other partitionings of the space. Two data structures stand out throughout the broad-phase, namely, spatial hashing and spatial partitioning trees.

Narrow-Phase: For the objects that overlap in the broad-phase, a check is performed between the query and a single primitive object that represents an obstacle. Often objects correspond to rigid bodies."

In the narrow-phase some very useful algorithms stand out when we take into account that most of the structure is described by primitive objects such as spheres. These algorithms include checking a point q and radius r for collision, calculating the distance from q to the nearest point on an obstacle and Calculating the nearest point p on the obstacle to a query point q.

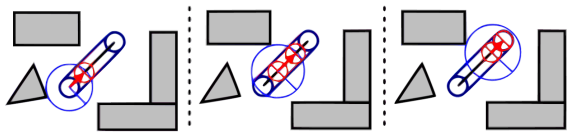

*Fig 2. Example swept circle obstacle check using only distance*

### 3.4: Optimized Spatial Hashing for Collision Detection of Deformable Objects[8]

In this paper the structure presented seems 'perfect' to work with GPU. The basic principle behind this structure is to create a spatial coherent hash table, meaning that the objects that are near in the data structure are also near in the three-dimensional space, resulting in efficient access and comparison between objects inside the table.

## 4. SPATIAL HASHING

Spatial hashing is a technique that is used to store elements in a hash table in which each key corresponds to a 3D coordinate and the value is a list of objects, in this particular case points, that are located in said space.

The basic principle is to split space into a variable number of cubes, where each cube can contain an arbitrary number of items. Additionally, any cube that does not contain an object, will not be created, being memory-friendly.

### 4.1: Operations of the data structure

*Hash function*: This operation will take the coordinates in meters of each point and divide it by the number required to create a 100 meter diagonal. It will then concatenate them and use this final string as a key.

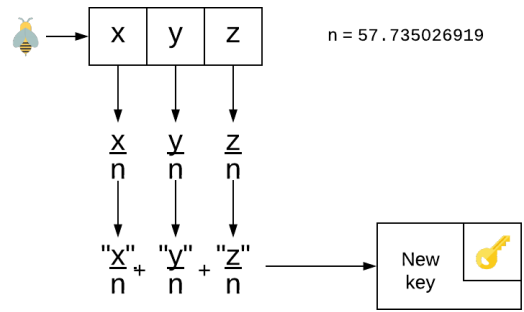

*Fig 3. Hash function*

*Create data structure*: This operation will receive the coordinates of a particular group of points from a file. The information provided will be used to generate a key, and proceed to insert each point in the position of the hash table that corresponds to said key. If it is the case that a point has already occupied this space then the point will be added to a doubly linked list of points.

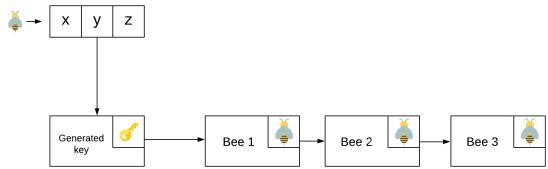

*Fig 4. Parse file*

*Concatenate list*: In order to keep track of the points at risk of collision we use an auxiliary linked list. This method will take said linked list and add the nodes of a second linked list to the end of it.

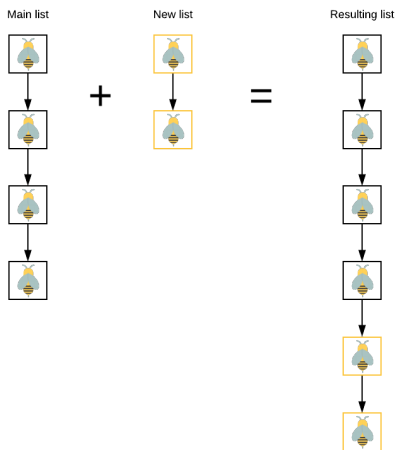

*Fig 5. Concatenate list*

*Find for single* point: Whenever in a cube is only one point it has to be compared with the points surrounding it in order to determine whether it is colliding or not. This method will take any point that is alone on a cube and, based on its coordinates, it will search all of the adjacent cubes in the spatial hash and compare the distance between all of the points contained in them and the current points.

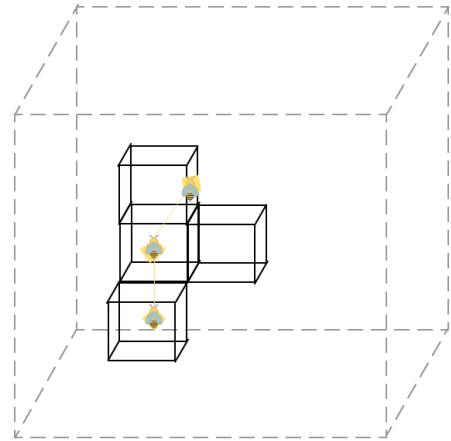

*Fig 6. Find for single* point

If it is the case that it is in risk of colliding it will be concatenated to our resultant list for colliding points. This process can be illustrated on figure X.

**4.2: Design criteria for the data structure**
Our data structure is required to represent the space intuitively, in order to improve the complexity of the *search collisions* method by decreasing the amount of operations performed to detect these events for a single point in space, but one that also is deterministic meaning we will find all the collisions existing on a given dataset.

Considering this, we decided spatial hashing would be a simple and yet efficient way to accurately represent our data. After analyzing various types of spatial data structures one could categorize them into two families: tree-based spatial structures (i.e. octrees, ball trees, KD-trees) and spatial hashing structures as the one we are presenting. We found tree-based spatial structures to have a worse complexity when it comes to deterministic collision detection. To determine whether a point is closer than a threshold using octrees would perform in O(n log n)[9]. While spatial hashing would take O(n) [10]. The algorithm used in octrees would be almost the same for all tree-based spatial structures, so the complexity would be the same. Also, access time for a leaf would be O(log n) in a balanced tree, opposed to O(1) amortized in a hash table. Moreover in our hash table we guarantee that there will be no collisions so O(1) is guaranteed.

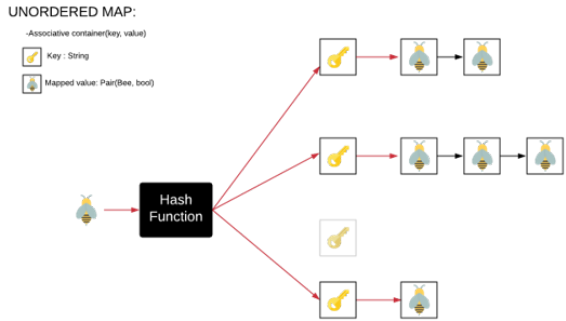

*Fig 7. Implementation of unordered map*

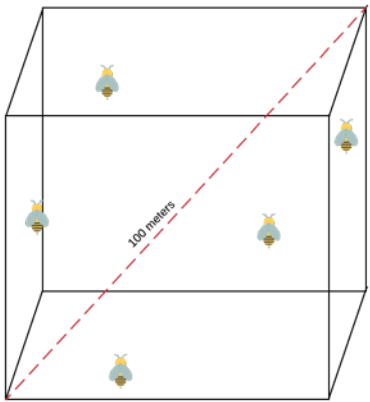

*Fig 8. Bucket structure*

### 4.3: Complexity analysis (e=space, n=number of points)

| Method | Complexity |
|---|---|
| Insert | O(1) |
| Find cube key | O(1) |
| Find for single point | O(e * n) |

### 4.4: Execution time and Memory used

The tests used two datasets from a very different nature. The ones with its size beginning with the digits "15" belong to a space about 8000 times bigger than the ones beginning with "10". In order to keep the results of the charts relevant we chose to only show two of the above described methods. Tests were ran on a PC with a AMD PRO A10-9700 R7 (2GHz) CPU and 8 GiB memory.

| Parse file (Parse file + hash function) | Memory (MiB) | Mean Time (ms) | Max Time (ms) | Min Time (ms) |
|---|---|---|---|---|
| **1500** | 27.7 | 2.721 | 3.14 | 0.193 |
| **100000** | 706 | 213 | 282 | 209 |
| **150000** | 454.5 | 281 | 391 | 276 |
| **1000000** | 3891.2 | 2106 | 2147 | 2068 |

| Find collision (Concatenate list + find for single point) | Memory (MiB) | Mean Time (ms) | Max Time (ms) | Min Time (ms) |
|---|---|---|---|---|
| **1500** | 32 | 0.199 | 0.314 | 0.193 |
| **100000** | 727 | 9 | 9 | 8 |
| **150000** | 487 | 51 | 52 | 51 |
| **1000000** | 4608 | 9 | 176 | 7 |

## 4.5: Result analysis

We can see from the third dataset that takes less memory than the previous one but takes longer finish. Knowing that the third one belongs to a much bigger space, and also having in mind that the points are uniformly distributed along the space we can conclude that the method *find for single point* is very expensive in terms of time when there are a considerable amount of of single points on the dataset, but when the bees are very likely to collide, the data structure performs very well in terms of time, but uses more memory.

## 5. CONCLUSION

After in depth analysis of spatial hashes as a solution to our problem, and describing relevant operations, by seeing the time results of the detect collisions method, it is clear that this solution scales nicely, as the difference in time between the datasets is rather small. Because of this, it is clear that this solution is much better than or first approach, which was a KD-tree, which worked fine for the smaller datasets, but scaled poorly.

## 5.1: Future work

In future work we would like to implement a similar data structure capable of handling data of moving bees, rather than a screenshot of their current position. Furthermore we would like to make our nearest neighbor algorithm more memory efficient by parallelizing the code. Also, future work should include a optimal hash function (e.g. using space-filling curves). Additionally, we would like to target a particular case in which our algorithm falters, which is when all 26 adjacent cubes on *find for single point* are empty bit of the following layer of adjacent cubes there is one in which a point at the same level as the point of the quearie might be colliding with it. Although this scenario is very unlikely for our particular problem, we would like to implement this update in order to generalize our solution.

## ACKNOWLEDGEMENTS

We would like to thank professors Pineda and Lalinde, from the Apolo Supercomputing Center.